\documentclass[runningheads]{llncs}
\usepackage{graphicx}
\usepackage{tikz}
\usepackage{comment}
\usepackage{amsmath,amssymb} % define this before the line numbering.
\usepackage{color}
\usepackage{capt-of}

\usepackage[accsupp]{axessibility}  % Improves PDF readability for those with disabilities.

\usepackage{makecell}
\usepackage{hyperref}
\hypersetup{
    colorlinks=true
    }

\begin{document}
% \linenumbers
\pagestyle{headings}
\mainmatter
\def\ECCVSubNumber{1103}  % Insert your submission number here

\title{Skeleton-free Pose Transfer for Stylized 3D Characters} % Replace with your title

\begin{comment}
\titlerunning{ECCV-22 submission ID \ECCVSubNumber} 
\authorrunning{ECCV-22 submission ID \ECCVSubNumber} 
\author{Anonymous ECCV submission}
\institute{Paper ID \ECCVSubNumber}
\end{comment}
%******************

% \begin{comment}
\titlerunning{Skeleton-free Pose Transfer for Stylized 3D Characters}
\author{Zhouyingcheng Liao$^{1}$\thanks{Work done during an internship at Adobe.},
Jimei Yang$^{2}$,
Jun Saito$^{2}$, \\
Gerard Pons-Moll$^{3,4}$, and
Yang Zhou$^{2}$ \\
}
\institute{$^1$Saarland University \,\,\,\,\,\,\,\, $^2$Adobe Research \,\,\,\,\,\,\,\, $^3$University of Tübingen \\ $^4$Max Planck Institute for Informatics, Saarland Informatics Campus}
\authorrunning{Z. Liao et al.}
% \institute{
% Saarland University \and Adobe Research \and University of Tübingen
% }
% https://github.com
% \end{comment}
%******************
\maketitle

% Custom commands
\newcommand{\R}{\mathbb{R}} % real number set
\newcommand{\verts}{\mathbf{V}} % vertices
\newcommand{\vertsrest}{\bar{\mathbf{V}}} % vertices
\newcommand{\vertex}{\mathbf{V}} % vertex
\newcommand{\face}{\mathbf{F}}  % mesh faces
\newcommand{\W}{\mathbf{W}}  % skinning weight matrix
\newcommand{\M}{\mathcal{M}} % mesh
\newcommand{\T}{\mathbf{T}}  % rigid transformation
\newcommand{\lowt}{\mathbf{T}}  % rigid transformation per row
\newcommand{\SE}{\mathbf{S}}  % shape embedding
\newcommand{\featy}{\mathbf{Y}} % intermediate feature (dummy)
\newcommand{\feat}{\mathbf{Z}} % intermediate feature
\newcommand{\centr}{\mathbf{C}} % center of the body part
\newcommand{\param}{\mathbf{\Theta}} % network parameters
\newcommand{\net}{\mathbf{\phi}} % GMEdgeNet
\newcommand{\rest}[1]{\bar{#1}}

\newcommand{\cmt}[1]{\textcolor{red}{[#1]}}

\definecolor{YangColor}{rgb}{0.7,0,0} % color for Yang
\newcommand{\yang}[1]{{\color{YangColor} \textbf{[Yang: #1]}}}
\newcommand{\revise}[1]{{#1}}

\definecolor{JimeiColor}{rgb}{0,0.6,0}
\newcommand{\jimei}[1]{{\color{JimeiColor} \textbf{[Jimei: #1]}}}

\definecolor{ZhouColor}{rgb}{0,0,0.8}
\newcommand{\zhou}[1]{{\color{ZhouColor} \textbf{[Zhou: #1]}}}

\definecolor{GerardColor}{rgb}{0.1,0.6,0.2} 
\newcommand{\GPM}[1]{{\color{GerardColor} \textbf{[Gerard: #1]}}}

\definecolor{JunColor}{rgb}{0.9,0.44,0.0} 
\newcommand{\jun}[1]{{\color{JunColor} \textbf{[Jun: #1]}}}

\centerline{\url{https://zycliao.github.io/sfpt}}

\begin{figure}
\vspace{-0.6cm}
\centering
\includegraphics[width=0.95\textwidth]{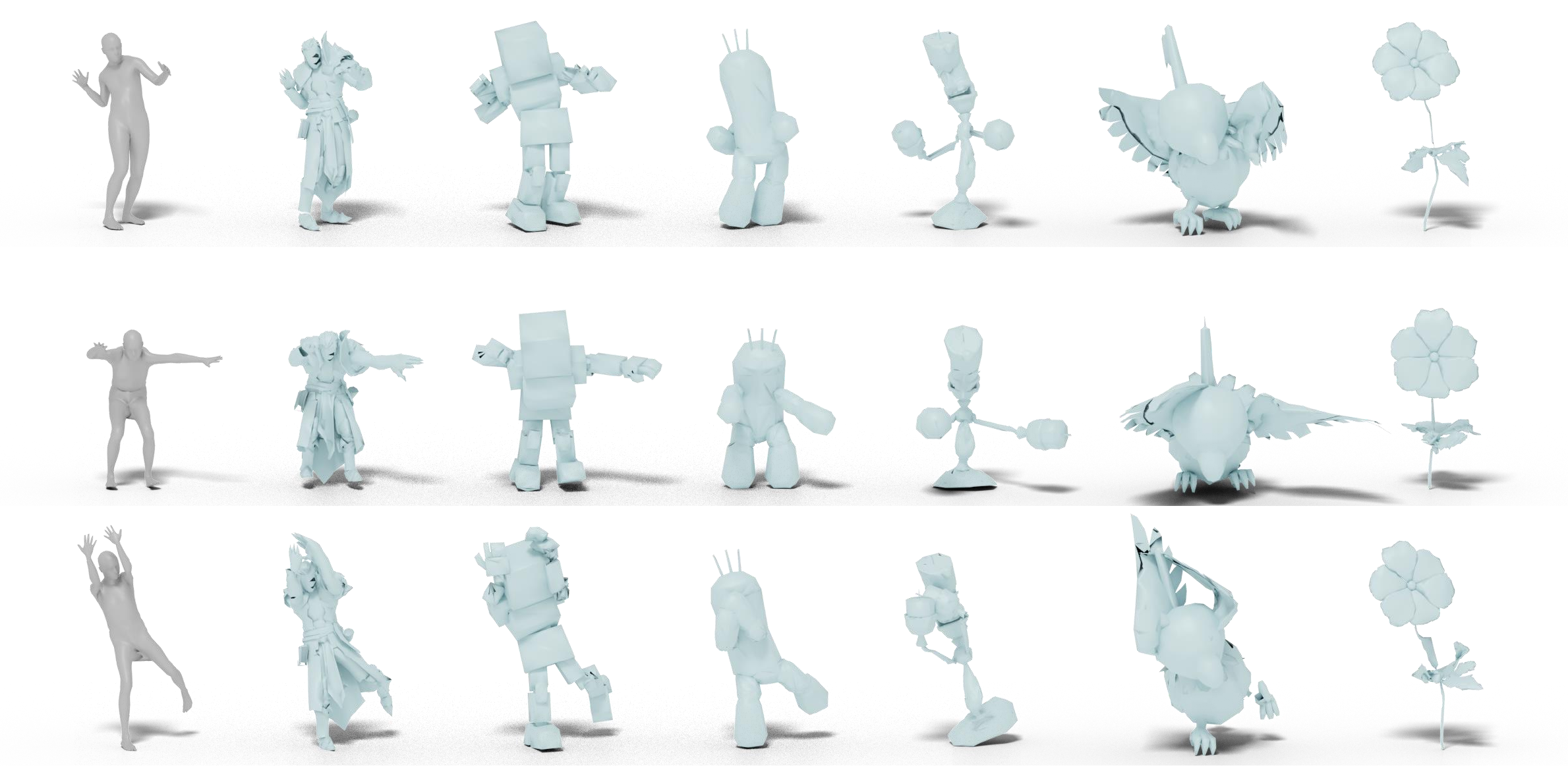}
\captionof{figure}
{Stylized 3D characters pose transfer. Given source pose characters as input (left), our model automatically transfers their poses to target subject characters with different body proportions and topologies (right). Our method does not require rigging, skinning, or correspondence labeling for both source and target characters.}
\vspace{-0.6cm}
\label{fig:teaser}
\end{figure}

\begin{abstract}
We present the first method that automatically transfers poses between stylized 3D characters without skeletal rigging. In contrast to previous attempts to learn pose transformations on fixed or topology-equivalent skeleton templates, our method focuses on a novel scenario to handle skeleton-free characters with diverse shapes, topologies, and mesh connectivities.  
The key idea of our method is to represent the characters in a unified articulation model so that the pose can be transferred through the correspondent parts. 
To achieve this, we propose a novel pose transfer network that predicts the character skinning weights and deformation transformations jointly to articulate the target character to match the desired pose. 
Our method is trained in a semi-supervised manner absorbing all existing character data with paired/unpaired poses and stylized shapes. It generalizes well to unseen stylized characters and inanimate objects. We conduct extensive experiments and demonstrate the effectiveness of our method on this novel task.

% \keywords{Character Pose Transfer, Auto-skinning, Correspondence Learning, 3D Deep Learning}
\end{abstract}

\section{Introduction}
Humans and animals evolved naturally with intrinsic articulation structures to facilitate their movements in complex environments. As a result, the articulated poses become an important part of their behaviors and emotions.
Based on this observation, 3D artists create highly-stylized characters in movies and games, from human-like, anthropomorphic
to even inanimate objects (e.g. Pixar's Luxo Jr.).
Posing and animating these characters like humans is key to conveying human-understandable expressions and emotions but actually requires many costly and sophisticated manual processes.
Furthermore, once animations are created, artists often want to re-use them in novel characters and scenarios.
A large number of existing studies have addressed the problem of automatically transferring poses between human-like characters~\cite{villegas2018neural,villegas2021contact,aberman2020skeleton,lee1999hierarchical,gleicher1998retargetting}, as they often share a similar or same skeletal rig.
Very few works were devoted to animating non-human characters from human data~\cite{gao2018automatic,baran2009semantic}, but they either require correspondence, or need to be trained for every pair of shapes.   
% hey gerard how about "examples" instead of keyframes? maybe. I basically don't get hte transition form keyframes to the limitation of requiring rigging and skeletons. Next paragraphs
% Baran requires correspondence. I am checking gao. 
% I think there is a better reference to clarify the previous work using example poses. I checked and Bharan rquires correspondence, and gao does not generalize to multiple subjects. Should we say that instead? fits better the flow of the intro. 
% The following part should talk more about the need of correspondence - let me check. Can you come to Slack? there is chat in overleaf. Maybe we chat there?
In this paper, we propose a learning-based method that automatically transfers poses between characters of various proportion and topology without skeletal rigs as a prerequisite (Fig.~\ref{fig:teaser}).
%
%Animating 3D characters is a fundamental task in many computer vision and computer graphics applications. 
%A common practice to animate a 3D character is to transfer the motion from another subject (e.g., motion capture data) to it. This usually needs to manually annotate the correspondence between the source and target character, or provides them with the same skeletal rigging \cite{x}. However, these requirements are costly and sometimes are not straightforward. It is impossible to assign a same set of skeletons for a person and an octopus. 
% In this paper, we propose a 3D character pose transfer method, which does not require rigging, skinning or any correspondence between characters. Given a posed character, our model can transfer its pose to other characters with different shapes and topologies (See Fig.~\ref{fig:teaser}. 

Given that most articulated 3D characters have piecewise rigid structures, the animation of characters is usually controlled by a set of sparse deformation primitives (rigs)~\cite{jacobson2014skinning} instead of manipulating the 3D mesh directly. 
%
% Among the popular rigging systems such as cage~\cite{ju2005mean} and deformation graph~\cite{sumner2007embedded}, the hierarchical skeleton is most commonly used.\jun{How about take this sentence out? I do not agree with this sentence, and also is distracting in explaining the pipeline with skinning by deformation primitives.}
%
With the deformation primitives, one has to bind them to the character mesh to build the correspondences between the primitives and the mesh vertices. This process is referred to as skinning. The skinning weight is used to describe how each deformation primitive affects the mesh deformation. 

For characters deformed by sparse primitives, transferring poses between them faces two challenges. 
First, rigging and skinning is a laborious manual process, which requires high expertise. 
Second, rigs must have correspondence to each other so that the primitives are paired to enable pose transfer from source to target characters.
Most existing works require the rigs to be exactly the same~\cite{villegas2018neural,wang2020neural,villegas2021contact}, e.g., a pre-defined human skeleton template.
However, in practice, the rig definition is arbitrary and the rig topology could differ a lot~\cite{xu2020rignet}. 
%
% It requires manual correspondence annotation and topology adjustment between the source and target rig. 
%
Thus, most existing pose transfer methods \cite{villegas2018neural,villegas2021contact,aberman2020skeleton,li2021nbs} cannot be directly applied to a new character without a rig or with a rig in a different topology.
While recent works~\cite{xu2020rignet,neuralskinning} achieve automatic rigging and skinning for characters, their output, i.e., hierarchical skeletons, do not have correspondence across different characters and cannot be directly used for pose transfer.

To address the above issues, we propose a novel pose transfer method to predict correspondence-preserving deformation primitives, skinning weights, and rigid transformations to jointly deform the target character so that its pose is similar to the one of the source character.

Specifically, we define a character articulation model in terms of a set of deformation body parts but without commonly-used hierarchical structures. 
%The deformation body parts share the unified notions and correspondences across characters (See Fig.~\ref{x}). 
%
Each part can be deformed by its own rigid transformation based on skinning weights. 
Besides, since there is no hierarchical structure to connect the body parts, they can be associated with any part of the character regardless of the shape topology. 
Given such an articulation model, we propose a novel deep learning based method to transfer poses between characters. It first generates skinning weights for both source and target characters, leading to a consistent segmentation of both into a set of corresponding deformation body parts. It then predicts a rigid transformation for each deformation body part. 
Finally, linear blending skinning (LBS)~\cite{Kavan2014-sm} is applied to deform the target mesh based on predicted rigid transformations. 
%The predicted rigid transformation is independent of both source and target character geometry, which is robust and adaptive to stylized characters.
As the deformation body parts and their rigid transformations are automatically generated by the network by analyzing the source and target meshes, they tend to be robust and adaptive to very stylized characters.

% We make the whole framework differentiable so that by imposing supervision on the mesh, the skinning and transformation can be learned simultaneously. Through our ablation studies, as the semantic-consistent skinning weight provides the semantic label for each mesh vertex, it helps the pose transfer network better encode the input mesh and achieve overall better pose transfer result. 
% First, it works as an attention map for the pose transfer network. Second, we use it to calculate the transformation of each skinning part between the pose mesh and its rest mesh as an input pose representation.

A lack of diverse data makes it hard to train our network. Most public character datasets with ground truth pose transfer data only contain natural human shapes~\cite{AMASS:ICCV:2019,ionescu2013human3,zhang2022couch} or characters of limited shape varieties~\cite{mixamo_2022}. The datasets with stylized characters~\cite{xu2020rignet} contain a single mesh in the rest pose for each character.
%since animating them is an arduous task.
We propose a semi-supervised training mechanism based on cycle consistency~\cite{zhu2017unpaired} to ease the requirement of ground truth pose transfer data on the stylized characters. This makes it possible to use arbitrary static characters with various shapes and topologies in training to improve the robustness of the method.
Overall, our contributions can be summarized as follows: 
% is a novel framework for rig-free 3D character pose transfer. Our method has the following advantages over existing methods:
\begin{enumerate}
    \item We propose the first automatic method for pose transfer between rig-free 3D characters with diverse shapes, topologies, and mesh connectivities.
    
    \item Our method parameterises the character pose as a set of learned independent body part deformations coherent across characters. We do not require any manual intervention or preprocessing, e.g., rigging, skinning, or correspondence labeling.
  
    % \item Our method predicts the semantic correspondence between characters and thus can support pose transfer between diverse shapes, topologies, and mesh connectivities.
  
     \item Our model is trained end-to-end and in a semi-supervised manner. We do not require neither annotations nor mesh correspondences for training and can make use of large amounts of static characters.
     
    %  \jun{It seems like the contribution boils down to the 1st one. 2nd and 3rd are repetitive and more like details of the 1st}
\end{enumerate}

\section{Related Work}
\subsubsection{Skeleton-based Pose Transfer.}

%Transferring pose based on skeletal rigs is a fundamental problem in computer graphics and animation. More than two decades ago,
Transferring poses based on skeletal rigs was intensively studied in the past.
Gleicher~\cite{gleicher1998retargetting} pioneered skeleton-based pose transfer through space-time optimization. Follow-up work~\cite{lee1999hierarchical,tak2005physically,al2018robust,choi2000online,aristidou2011fabrik} mostly incorporated various physics and kinematics constraints into this framework. Generalization to arbitrary objects has been proposed by \cite{yamane2010animating,Rhodin2015-bc}, but they require example poses from users.
Recent deep learning methods~\cite{villegas2018neural,villegas2021contact,lim2019pmnet} trained neural networks with forward kinematics layers to directly estimate hierarchical transformations of the target skeleton. However, their models require the source and target skeletons to be the same while only allowing for different proportions. 
\cite{baran2007pinocchio,li2021nbs} fit a predefined template skeleton and derive the skinning weights for character pose transfer within the same category, e.g., humanoid, quadruped, etc. \cite{poirier2009rig} relaxed the singular template constraint through a multi-resolution topology graph. \cite{aberman2020skeleton} proposed skeleton-aware pooling operators which supports skeletal pose transfer with a different number of joints. Yet, these methods still require the skeletons to be topologically equivalent. However, even these relaxed skeleton constraints cannot be guaranteed through state-of-the-art automated rigging and skinning methods~\cite{xu2020rignet,xu2019predicting,neuralskinning}.
Our method does not require skeletal rigging and can transfer poses across characters with different topologies. 

\subsubsection{Mesh Deformation Transfer.}

Character pose transfer can also be achieved by mesh deformation transfer without relying on rigs. Traditional methods~\cite{sumner2004deformation,ben2009spatial,baran2009semantic,avril2016animation} require accurate correspondences through manual annotation.
\cite{yang2018biharmonic} proposed a keypoint detection method to characterize the deformation, but still required user effort to specify corresponding keypoints from the source to target. 
Recent deep learning based methods analyzed the mesh deformation primitives and embedded the mesh into latent spaces~\cite{tan2018variational,tan2018variational,yang2020multiscale}.
% Recent deep learning based methods embedded the mesh into a latent space and analyze deformation primitives~\cite{tan2018variational,tan2018variational,yang2020multiscale}. \jimei{What does it mean by "analyzing deformation primitives?}
%
However, their latent spaces are not shared across subjects and thus cannot be used for pose transfer between different characters.
% However, they are generative models and cannot be applied to pose transfer tasks. \jimei{Why?} 
%
\cite{gao2018automatic} trained a GAN-based model with unpaired data in two shape sets to perform pose transfer. But it is limited to shape deformation in the training set and does not generalize to unseen characters. 
% \jimei{check this}. \yang{yes: they need to train on two sets of shapes from both characters, then apply pose transfer. No generalization }
%
\cite{wang2020neural,zhou2020unsupervised} disentangled the shape identity and pose information and made it possible to transfer poses to unseen characters. However, they can only handle characters with limited shape varieties, e.g, human bodies with minimal clothing. 
% Besides, their network structures~\cite{bouritsas2019neural,wang2020neural} require fixed mesh connectivity, which significantly limits the applicability. 
Our method automatically generates consistent deformation body parts across different character meshes and deforms the target mesh with part-wise rigid transformations in an LBS manner. Hence, no manual correspondence is needed. Once trained, our network can generalize to unseen characters with various shapes and topologies.

% For the state-of-the-art 3D shape representation methods, i.e., implicit function and Nerf, reconstructing articulated meshes still need the rig as a proxy.
% arap, acap
% That is because the neural network can hardly reconstruct articulated characters of diverse shapes and preserve fine details\cmt{Zhou: anything to cite?}.  Although mesh deformation transfer methods might require less manual work than rig-based methods, they are constrained in different aspects. 

\subsubsection{Correspondence Learning.}

Correspondence learning is crucial in pose transfer and motion retargeting tasks~\cite{groueix20183d,aberman2020skeleton,jakab2018unsupervised,siarohin2019animating,siarohin2019first,song20213d,saito2021scanimate}. \cite{jakab2018unsupervised,siarohin2019animating,siarohin2019first} detected 2D keypoints on images as correspondences for human video reposing. \cite{hung2019scops,siarohin2020motion2,siarohin2021motion} found corresponding regions and segments for pose transfer. \cite{reed2015deep} performed analogies on 2D character sprites and transferred animations between them. They worked on image domain by utilizing deep image features to locate corresponding body parts. \cite{liu2021deepmetahandles,shi2021skeleton,jakab2021keypointdeformer} proposed unsupervised methods to discover corresponding 3D keypoints as deformation handles. They generate plausible shapes, e.g., chairs, airplanes, via shape deformation but are not suitable for character posture articulation. \cite{song20213d} found per-vertex correspondence between human meshes via correlation matrices. But its generalization is limited to shapes close to training data~\cite{bhatnagar2019multi,bogo2014faust}.
Our method discovers part-level shape correspondence by learning through the pose transfer task. It does not need correspondence annotation for supervision and can generalize to unseen stylized characters.

\section{Method}
% Our goal is to transfer the pose of a character to another character. These characters are rig-free, which means their rigging and skinning are not required. Their topology can be different and we do not require any correspondence between them.
% To this end, we propose a neural model which learns shape correspondence and pose transfer simultaneously. Our key assumption is that for 3D articulated characters, learning shape correspondence and learning pose transfer can facilitate each other.

% \subsubsection{Conventions.}
% We represent matrices with bold capital letters (e.g., $\face$). Lower-case letters in bold represent vectors (e.g., $\vertex$). Special Latin letters denote sets (e.g., $\M$). Normal fonts denote scalars or functions (e.g., $w, f$). We use the superscript to denote the subject of a character (e.g., $\verts^s$). The bar symbol indicates the character is the rest pose (e.g., $\rest{\verts}$). The hat symbol denotes the variable is predicted by our model (e.g., $\hat{\T}$).

%--------------------------------------------------Overview--------------------------------------------------
\subsection{Overview}

\begin{figure}[t]
    \centering
    \includegraphics[width=0.95\textwidth]{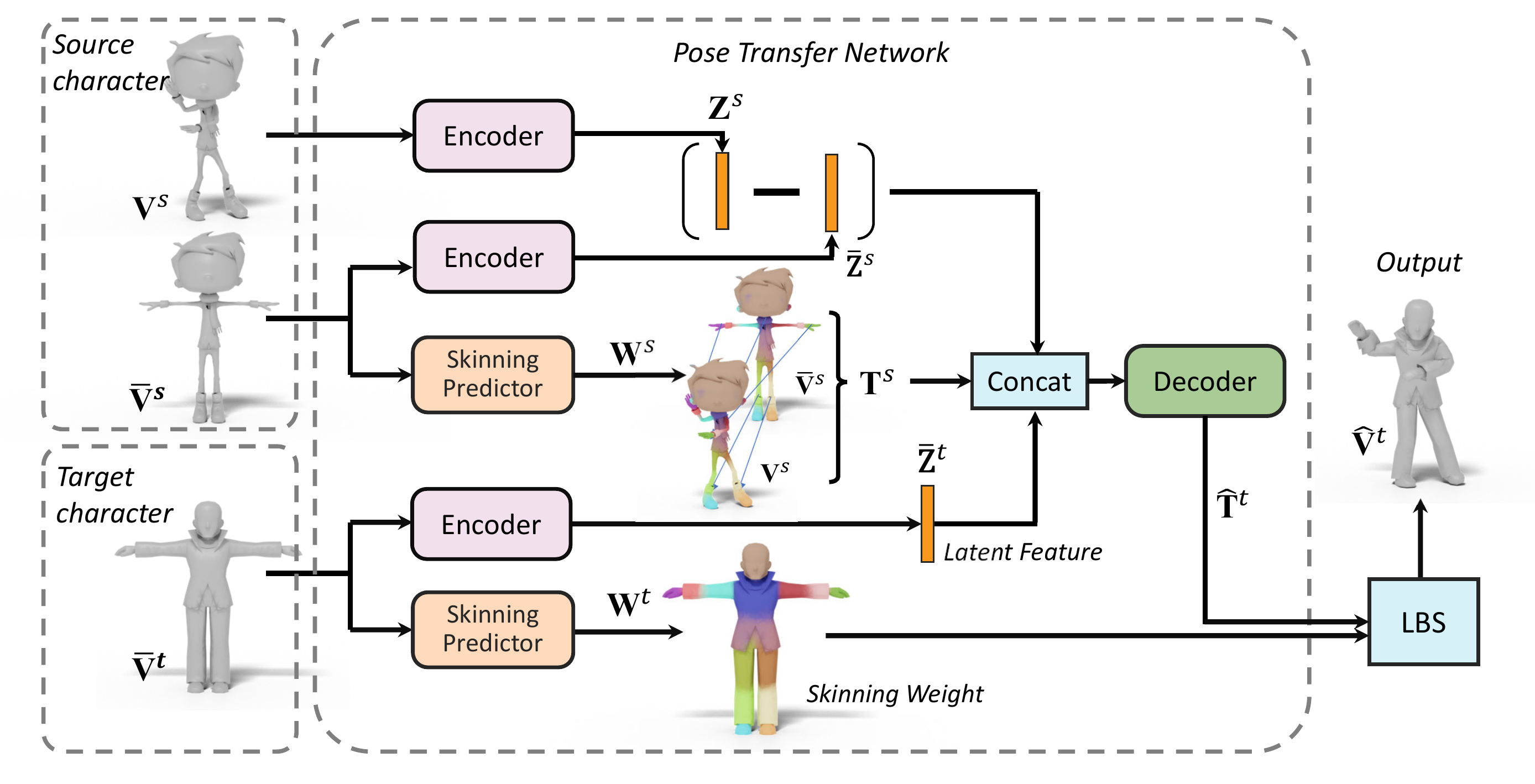}
    \caption{Overview. Given a posed source character and a target character as input, the pose transfer network estimates character skinning weights and part-wise transformations which articulate the target character through LBS to match the pose of the source.}
    \label{fig:network}
\end{figure}

Given a source 3D character mesh $\verts^s$ with the desired pose and its mesh $\vertsrest^s$ in rest pose 
%\jun{and in its rest pose $\vertsrest^p$ too?} 
and a different target 3D character mesh $\vertsrest^t$ in rest pose,
the goal of our method is to deform the target mesh to a new pose $\hat{\verts}^t$ which matches the input source pose $\{\verts^s, \vertsrest^s, \vertsrest^t\} \mapsto \hat{\verts}^t$. Here, we use the bar symbol $\vertsrest$ to indicate the character in rest pose. To solve this problem, we propose an end-to-end neural network that learns part-level mesh correspondences and transformations between characters to achieve pose transfer. The overview is shown in Fig.~\ref{fig:network}.

We first define a character articulation model to represent the mesh deformation in terms of a set of deformation parts (Sec.~\ref{sec:deformation}). Unlike existing methods~\cite{aberman2020skeleton,baran2007pinocchio,villegas2018neural} requiring skeletal rigging to deform character body parts hierarchically, our model deforms body parts independently without the kinematic chain. Our method parameterises the character pose as a set of learned independent body part deformations coherent across characters, which is the foundation for the following pose transfer network to overcome topology constraints.

% \jimei{After reading this section, I feel it may be better to organize it as follows: 1) introduce the pose transfer network and 2) the pose transfer network consists of several modules: skinning weight predictor, (shape/pose) encoder and transformation decoder.}
% \yang{Revised based on Jimei's comment}
We propose a novel skeleton-free pose transfer network to predict the skinning weights and the transformations for each deformation part defined in the above character articulation model so that the target character can be transformed by linear blending skinning (LBS) to match the input pose (Sec.~\ref{sec:pose-transfer}).

The pose transfer network consists of three modules: \textit{skinning weight predictor}, \textit{mesh encoder}, and \textit{transformation decoder}. The skinning weight predictor estimates per-vertex skinning weights that segment the mesh into $K$ deformation parts (see examples in Fig.~\ref{fig:deformation_part}).
% \jimei{We need to define well the terms "primitive, segment, part" to be consistent with the remaining sections.} \yang{sure, revised them.}
The \textit{mesh encoder} encodes the input mesh into a latent feature that embeds both pose and shape information. The \textit{transformation decoder} predicts a set of part-wise rigid transformations, which are further used to articulate the target mesh into the desired pose.

% With predicted mesh correspondence, we propose a novel \textit{pose transfer network} to predict the transformation matrix for each corresponding part and apply LBS to generate the final reposed character (Sec.~\ref{sec:pose-transfer}). Specifically, we design an encoder network to encode the pose and shape of the input meshes into latent vectors and an decoder network takes as input such vectors and predict part-specific transformations.\zhou{A little confusing. }

We train our framework end-to-end in a semi-supervised manner (Sec.~\ref{sec:training}). For characters with pairwise animation sequences~\cite{AMASS:ICCV:2019,mixamo_2022}, i.e. different subjects with the same animation poses, we train our network directly with cross-subject reconstruction. There also exist datasets with stylized characters of diverse shapes, topologies, and mesh connectivities. However, such data usually contains only a static rest pose and thus cannot be directly used in training. We propose a cycle-consistency loss to train on such data unsupervised, which turns out to improve our model robustness significantly.

\subsection{Character Articulation Model}
\label{sec:deformation}
\begin{figure}[t]
    \centering
    \includegraphics[width=0.9\textwidth]{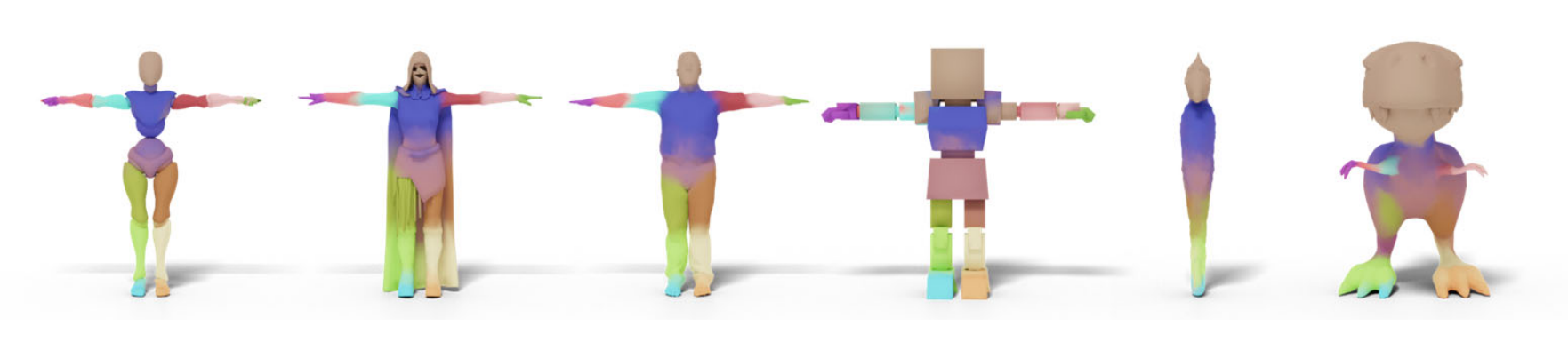}
    \caption{Visualization of deformation parts based on predicted skinning weights. Each color represents a deformation part. The deformation part is semantically consistent across characters with various shapes and topologies.}
    \label{fig:deformation_part}
\end{figure}

We propose to represent the mesh deformation in a unified way so that the pose can be easily transferred between characters  with various shapes and topologies.
We define $K$ deformation parts for a mesh $\vertsrest$ with $N$ vertices. Each part can be deformed based on the skinning weight $\W \in \R^{N \times K}$ associated with it.
The K deformation parts are not character-specific but consistent across characters (Fig.~\ref{fig:deformation_part}).
% \jimei{Better not mention semantics. We could say: "share unified notions across characters. A figure will be helpful with colored part segmentation of different characters.} \yang{agreed}.
$\W$ satisfies the partition of unity condition where $0 \leq w_{i,k} \leq 1$ and $\sum_{k=1}^K w_{i,k} = 1$. Here $i$ is the vertex index and $k$ is the deformation part index.
Different characters may have different shapes and topologies, so ideally the number of deformation parts should vary. We define $K=40$ as the maximum number of parts for all the characters in our experiment. Depending on the shape of the character, we allow some parts to be degenerate, i.e. having zero coverage: $w_{i,k} = 0, \forall i$. Meanwhile, the number of vertices $N$ is not fixed and can vary from character to character during either training or testing phases.

Given rigid transformations for $K$ body parts $\T=\{ \lowt_1, ..., \lowt_K  \}$, we use LBS~\cite{Kavan2014-sm} to deform the character mesh,
\begin{equation}
    \vertex_i = \sum_{k=1}^K w_{i,k}\T_k (\vertsrest_{i} - \centr_k),\ \ \ \forall \vertsrest_i \in \vertsrest
\end{equation}
where $\centr_k$ is 
% the rotation center of each body part. Its position is 
the center of deformation part $k$ in terms of the average position of vertices weighted by the skinning weight,
\begin{equation}
    \centr_k =  \frac{\sum_{i=1}^{N} w_{i,k}\vertsrest_i}{\sum_{i=1}^{N} w_{i,k}} 
\end{equation}

% Unlike the common skeleton-based skinning, we do not connect bones to handle characters with various shapes and topologies. Thus, our network predicts the transformation $\T$ directly without the kinematic chain.
We do not connect the center of deformations parts $\centr_k$ to form a skeleton since the skeleton connectivity varies in characters with different topology~\cite{xu2019predicting}. Our transformation $\lowt_k$ is applied independently on each deformation part without the kinematic chain.
A consistent deformation part segmentation together with part-wise transformations forms a more general way of articulation than the commonly-used skeleton-based methods~\cite{aberman2020skeleton}, which is crucial for achieving the skeleton-free pose transfer.

%--------------------------------------------------Pose transfer Network --------------------------------------------------

\subsection{Skeleton-Free Pose Transfer Network}
\label{sec:pose-transfer}

We propose a skeleton-free pose transfer network to transfer the pose from a source character to a different target character (see Fig.~\ref{fig:network}). It predicts skinning weights of both source and target characters through the skinning weight predictor and estimates the corresponding transformation matrices jointly from the mesh encoder and transformation decoder network.

%--------------------------------------------------Skinning predictor--------------------------------------------------
\subsubsection{Skinning Weight Predictor}
\label{sec:skinning-predictor}
Given a character mesh $\vertsrest$, we design a graph convolution network $g_s$ to predict its skinning weight $\W \in \R^{N \times K}$,
\begin{equation}
    \W = g_s(f(\vertsrest); \net_s)
\end{equation}
where $f(\vertsrest)\in \R^{N \times 6}$ is the vertex feature vector consisting of position and normal. $\net_s$ are learnable parameters. Each row of $\W$ indicates each vertex skinning weight association to $K$ deformation parts. 
The network architecture follows \cite{xu2020rignet} and can process meshes with arbitrary number of vertices and connectivities. We modify the last layer as a softmax layer to satisfy the skinning weight convex condition. The detailed structure can be found in the supplementary.

%--------------------------------------------------Pose transfer--------------------------------------------------

\subsubsection{Mesh Encoder.}
We use another graph convolution network $g_e$ to encode the input mesh $\verts$ into a latent feature $\featy \in \R^{N \times C}$,
\begin{equation}
    \featy = g_e(f(\verts);\net_e) 
\end{equation}
where $C$ is the dimension of the latent space and $\net_e$ are learnable parameters.

Instead of pooling $\featy$ into a global latent feature, we multiply it with the predicted skinning weight as an attention map to convert the feature dimension $\R^{N \times C} \mapsto \R^{K \times C}$. This conversion can be interpreted as an aggregation function to gather deformation part features from relevant vertices. After that, a 1D convolution layer is further applied to transform the feature to be the attended latent feature $\feat \in \R^{K \times C}$,
\begin{equation}
    \feat = \text{Conv1d}(\W^\intercal \cdot \featy, \net_c )
\end{equation}
where $\net_c$ are learnable parameters and $C=128$ in our experiment. 
Note that the mesh encoder is applied to all three input meshes $\verts^s, \vertsrest^s, \vertsrest^t$ to obtain their attended latent features ${\feat}^{s}, \bar{\feat}^{s}, \bar{\feat}^{t}$ with corresponding skinning weights.

% \jimei{How does it correspond to shape and pose latent features?} \yang{$\feat$ is our latent feature}
% We do not let gradients from the shape embedding backpropagated to $f_s$, because these gradients will affect the deformation quality of the skinning weight.

\subsubsection{Transformation Decoder.}
The goal of the decoder is to predict transformations $\hat{\T}^t=\{ \hat{\lowt}^t_1, ..., \hat{\lowt}^t_K  \}$ on each deformation part of the target mesh $\vertsrest^t$. Hence, the target mesh $\vertsrest^t$ can be reposed to $\hat{\verts}^t$ which matches the desired pose mesh $\verts^s$. The decoder takes as input three component: 

\begin{itemize}
    \item the latent feature of the target mesh $\bar{\feat}^t$. It is derived from the target mesh $\vertsrest^t$ with the mesh encoder. It encodes the target mesh shape information.
    
    \item the difference between the latent features of the posed source mesh and itself in rest pose ${\feat}^{s}-\bar{\feat}^{s}$. %${\feat}^{s}$ is derived from the desired posed mesh $\verts^s$ and $\bar{\feat}^{s}$ from ${\vertsrest}^{s}$ in rest pose. 
    %We found that directly injecting the latent feature $\feat^s$ returns postures with less articulation and thus are less accurate.
    
    \item the transformation of each deformation part $\T^s = \{\lowt^s_1, ..., \lowt^s_K\}$ between the pair of source meshes. This explicit transformation serves as an initial guess and helps the network focus on estimating residuals. It is analytically calculated by \cite{besl1992method}.
\end{itemize}

% The ablation study shows the efficiency of each input. 
To summarize, the decoder takes the concatenation of $\feat^t, {\feat}^{s}-\bar{\feat}^{s}, \T^s$ as input and predicts the transformation $\hat{\T}^t$:
\begin{equation}
    \hat{\T}^t = g_d(\bar{\feat}^t, {\feat}^{s}-\bar{\feat}^{s}, \T^s; \net_d)
\end{equation}
where $\net_d$ are learnable parameters for the decoder. With the predicted skinning weights $\W^t$ and transformation $\hat{\T}^t$, we can use the proposed articulation model to deform the target mesh $\vertsrest^t$ to the new pose $\hat{\verts}^t$.

% The goal of the decoder is to process shape information from the subject mesh and pose information from the pose mesh and predict proper transformations for the subject mesh.
% We use the embedding of the subject mesh $\rest{\SE}^p$ as its shape representation. For pose information, we first use the difference of the pose mesh embedding and its rest mesh  embedding $\SE^p - \rest{\SE}^p$. 
% To make the network better aware of the input pose, we add the transformation of the pose mesh $\T^p$ as an additional input. Given the predicted skinning weight of the pose mesh $\W^p$, $\T^p$ can be analytically calculated.
% The decoder takes the concatenation of $\rest{\SE}^s, \SE^p - \rest{\SE}^p, \T^p$ as input and predicts the transformation $\hat{\T}^s$:
% \begin{equation}
%     \hat{\T}^s = f_d(\SE^p - \rest{\SE}^p, \rest{\SE}^s, \T^p)
% \end{equation}

%--------------------------------------------------Training--------------------------------------------------
% \vspace{-3mm}
\subsection{Training and Losses}
\label{sec:training}

We propose the following losses to train our network in a semi-supervised manner to make the best use of all possible data. 

% The articulation model and skinning weight predictor is trained without direct supervision but trained end-to-end with the pose transfer task. Details are shown as follows.

\subsubsection{Mesh reconstruction loss.}
For characters with paired pose data
% \jimei{Not clear} \yang{explained above}
in \cite{AMASS:ICCV:2019,mixamo_2022}, we use a reconstruction loss as direct supervision. 
% Given a source pose mesh $\verts^s$ and a target shape mesh $\verts^t$ in the rest pose, our model can predict the output mesh $\hat{\verts}^t$. The output mesh has the same mesh connectivity as the ground truth mesh $\rest{\verts}^s$.
We apply a per-vertex L1 loss between the predicted mesh $\hat{\verts}^t$ and the ground truth mesh $\verts^t$,
\begin{equation}
    L_{rec} = || \hat{\verts}^t - \verts^t ||_1
\end{equation}
% In addition to training with paired pose data, this loss can also be applied to a single character with multiple poses, where the input to the network $\verts^s$ and $\vertsrest^s$ are from the same character but with different poses. 
% \jimei{Not clear. Do we always need T-posed character as input? I change it to be consistent with the text above. Not sure what you're actually doing.} \yang{revised, but this could be put into supp since it's a variant loss we used for limit data}

\subsubsection{Transformation loss.}
With the predicted skinning weight $\W^t$ of the target mesh, we group the vertices into $K$ deformation parts by performing $\mathop{\mathrm{argmax}}_k {w_{i,k}}$. Then we calculate the ground truth transformation $\T^t$ on the approximated parts between the input rest pose mesh $\vertsrest^t$ and the ground truth mesh $\verts^t$. We apply an L1 loss between the ground truth and predicted transformation $\hat{\T}^t$,
\begin{equation}
    L_{trans} = || \hat{\T}^t - \T^t ||_1
\end{equation}
% In this loss, the subject of the pose mesh and the subject mesh can be the same. This is applied when paired data $\{\verts^p, \rest{\verts}^s, \verts^s\}$ is not available but the character has multiple poses. In this case, the pose mesh and the subject mesh are $\verts^s, \rest{\verts}^s$ and the target mesh is $\verts^s$.

\subsubsection{Cycle loss.}
When paired data are not available, or just a single rest pose mesh is provided, e.g., in \cite{xu2020rignet}, we use the cycle consistency loss for training~\cite{zhu2017unpaired}. Given a pair of source meshes $\verts^s, \vertsrest^s$ and a target mesh $\vertsrest^t$ in rest pose, we first transfer the pose from source to target: $\{\verts^s, \vertsrest^s, \vertsrest^t\} \mapsto \hat{\verts}^t$, and then transfer the pose from the predicted target back to the source mesh: $\{\hat{\verts}^t, \vertsrest^t, \vertsrest^s\} \mapsto \hat{\verts}^s$. The predicted source mesh $\hat{\verts}^s$ should be the same as $\verts^s$. We apply L1 loss between them for the cycle reconstruction.

% \begin{equation}
%     L_{cyc} = || \hat{\verts}^s - \verts^s ||_1
% \end{equation}

% This loss can enable the training on characters with a single rest pose. 

Through experiments, we found that training only with this loss leads to mode collapse. In existing datasets, characters with multiple poses are usually human characters, while most stylized characters are only in rest pose. These stylized characters can only be used as the target mesh $\vertsrest^t$ in the cycle loss instead of interchangeably as $\verts^s$. Thus the network tends to collapse and results in $\hat{\verts}^t$ with limited pose variance. 
To solve this problem, we apply the transformations calculated from the source meshes $\T^s$ to the target mesh $\vertsrest^t$ to obtain a pseudo-ground truth $\tilde{\verts}^{t}$ for $\hat{\verts}^t$. The complete cycle loss is
\begin{equation}
    L_{cyc} = || \hat{\verts}^s - \verts^s ||_1 + w_\text{pseudo} || \hat{\verts}^t - \tilde{\verts}^{t} ||_1
\end{equation}
where $w_{pseudo}=0.3$ is used in our experiment. To note that $\hat{\verts}^{t}$ is an approximated pseudo-ground truth mesh and sometimes may not be well-deformed if the transformation $\T^s$ is large. We introduce this term as a regularization which helps prevent the model from collapsing.

\subsubsection{Skinning weight loss.}
In existing rigged character datasets~\cite{AMASS:ICCV:2019,mixamo_2022,xu2020rignet}, the skeletons and skinning weights are defined independently for each character. 
Therefore, we cannot use them directly as ground truth to supervise the training of our skinning weight predictor because of their lack of consistency.
%Though we cannot use the ground truth skinning weight as direct supervision to train the skinning weight predictor, we can still make use of it in another way through contrastive learning~\cite{chen2020simple}.  
We thus propose a contrastive learning method to make use of such diverse skinning data.
Our assumption is if two vertices belong to the same body part based on the ground truth skinning weight, they should also belong to the same deformation part in the predicted skinning.
% Here, we define the vertex $i$ belongs to the body part $k$ if $w_{i,k} > 0.9$. 
% If the vertex $i$ does not have any skinning weight greater than $0.9$, i.e., $\max_j w_{i,k} \leq 0.9$, it is omitted in this loss. 
We select vertices with $w_{i,k} > 0.9,\ \exists k$ and use the KL divergence to enforce similarity between skinning weights of two vertices,
\begin{equation}
    L_{skin} = \gamma_{i,j} \sum_{k=1}^K (w_{i,k} log(w_{i, k}) -  w_{i,k} log(w_{j, k}))
\end{equation}
where $i,j$ indicate two randomly sampled vertices. $\gamma$ is an indicator function: $\gamma_{i,j}=1$ if vertices $i$ and $j$ belong to the same part in the ground truth skinning weight and $\gamma_{i,j}=-1$ if not. 
This loss holds only when the ground truth skinning is available.
% \yang{add mention its optional}

% If there is no ground truth skinning weight, to make the predicted skinning weight distinguishes different body parts, we impose a Chamfer loss between vertices and the predicted part centers, \yang{what is this loss?}
% \begin{equation}
%     L_{skin} = Chamfer(\centr, \verts)
% \end{equation}

\subsubsection{Edge length loss.}
The desired deformation should be locally rigid and preserve the character shape, e.g., limb lengths and other surface features. Thus, we apply an edge length loss between the predicted mesh and input target mesh to prevent undesired non-rigid deformations,
\begin{equation}
    L_{edge} = \sum_{\{i,j\} \in \mathcal{E}} | \ || \hat{\vertex}^t_i - \hat{\vertex}^t_j ||_2  - || \vertex^t_i - \vertex^t_j ||_2 \ |
\end{equation}
where $\mathcal{E}$ denotes the set of edges on the mesh.

\section{Experiments}

\subsection{Datasets}
We train our model on three datasets: AMASS~\cite{AMASS:ICCV:2019}, Mixamo~\cite{mixamo_2022} and RigNet~\cite{xu2020rignet}. We additionally use MGN~\cite{bhatnagar2019multi} for evaluation.

% \vspace{-5mm}
\subsubsection{AMASS~\cite{AMASS:ICCV:2019}} is a large human motion dataset that fits SMPL~\cite{loper2015smpl} to real-world human motion capture data. SMPL disentangles and parameterizes the human pose and shape. Therefore, we can obtain paired pose data by switching different shape parameters while keeping the pose parameter the same. We follow the train-test split protocol in \cite{zhou2020unsupervised}.

% \vspace{-5mm}
\subsubsection{Mixamo~\cite{mixamo_2022}} contains over a hundred humanoid characters and over two thousand corresponding motion sequences. Since the motion sequences are shared across all characters, the paired pose data is also available. We use 98 characters for training and 20 characters for testing. The detailed split can be found in the supplementary.

% \vspace{-5mm}
\subsubsection{RigNet~\cite{xu2020rignet}} contains 2703 rigged characters with a large shape variety, including humanoids, quadrupeds, birds, fish, etc. All characters have their skeletal rigging. Each character only has one mesh in the rest pose. 
% In this work, we mainly focus on characters that can be animated by human motion. 
We remove character categories that can not be animated by the human pose, e.g., fish. We follow the train-test split protocol in \cite{xu2020rignet} on the remaining $1268$ characters.

% \vspace{-5mm}
\subsubsection{MGN~\cite{bhatnagar2019multi}} contains 96 scanned clothed human registered to SMPL topology. We evaluation on this dataset to further demonstrate the robustness on human model with larger shape variety.

% \subsubsection{Preprocessing.}
% All meshes were rescaled to have the same height. We also translated the mesh to the center and frontalized the character. We used the mesh simplification and subdivision to make the mesh face number between 2000 and 5000. 

\subsection{Comparison Methods}
\textbf{Pinocchio}~\cite{baran2007pinocchio} is a pioneering work in automatic 3D character rigging and skinning. It fits a predefined skeleton to the target mesh and calculates skinning weights. As the skeleton is fixed, it achieves pose transfer by retargeting the joint rotations estimated by \cite{besl1992method}. Pinocchio has a strict requirement on the mesh connectivity: non-watertight, multi-component, or non-manifold meshes are not allowed. We manually preprocessed meshes to match its requirement in our evaluation. 
% It may fail when the shape and topology of the mesh do not match the pre-defined skeleton well. 
\textbf{Skeleton-aware Network (SAN)}~\cite{aberman2020skeleton} transfers the pose between two humanoid characters with the same skeleton topology. However, they require the motion statistics for both source and target characters to remap the predicted motion, e.g., an animation sample for the test subject mesh. This is not available in our task where only one instance of the subject character is provided. To make a fair comparison, we used the average statistics from the training set for test meshes. The ground truth skeleton and skinning is also assumed given for this method.
\textbf{Neural Blend Shape (NBS)}~\cite{li2021nbs} is the state-of-the-art method that achieves pose transfer between skeleton-free human shapes. They adopt SMPL skeleton template and can only work on human characters. 
\textbf{Shape Pose Disentanglement (SPD)}~\cite{zhou2020unsupervised} can disentangle pose and shape information for 3D meshes from the same category. The pose transfer can be achieved by applying the pose information from one mesh to the other while keeping the shape information. Their model can only work on meshes with the same connectivity and thus we evaluate it only on SMPL-based dataset.
\textbf{Ours (AMASS)} represents the reposing results from our proposed framework. It is trained only with AMASS data with limited character shapes. It is used for evaluating the generalization of the proposed network architecture. \textbf{Ours (full)} is our full result trained on all three datasets mentioned above.

\subsection{Pose Transfer Evaluation}

We evaluate our skeleton-free pose transfer results on different stylized characters both qualitatively and quantitatively. 

\begin{figure}[t]
    \centering
    \includegraphics[width=0.9\textwidth]{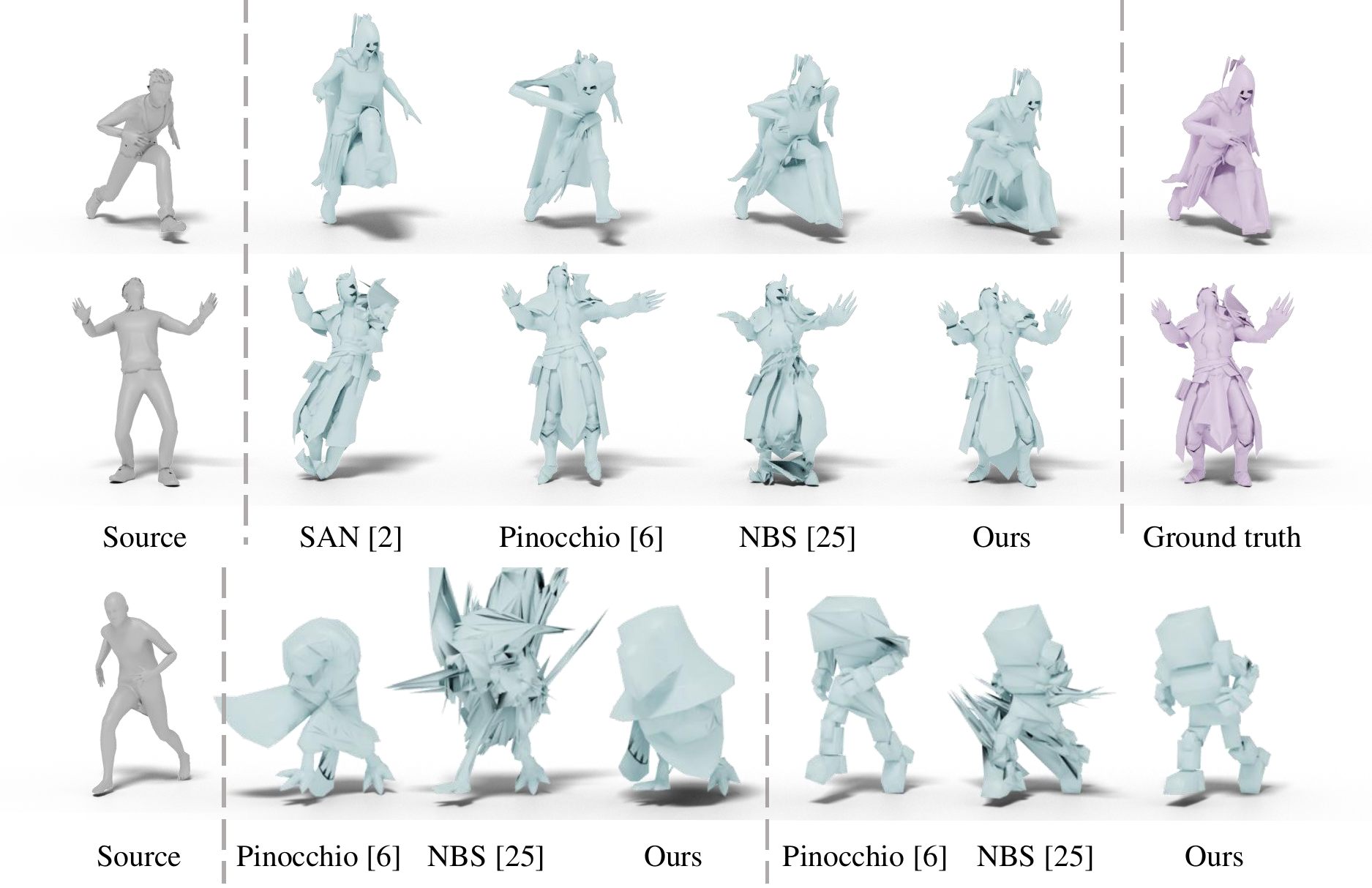}
    \caption{Pose transfer results for human (top) and stylized characters (bottom).}
    \label{fig:quali-comparison}
\end{figure}

Fig.~\ref{fig:quali-comparison} shows comparison results of the reposed characters from each of the comparison methods. SAN~\cite{aberman2020skeleton} fails on our task where only a single test mesh is given. It relies a lot on the motion statistics for test characters. Pinocchio~\cite{baran2007pinocchio} does not preserve the character shape well, e.g., the limbs have undesired non-rigid deformations. NBS~\cite{li2021nbs} results in collapsed shapes and cannot generalize well to stylized characters. 
% Its prediction collapses when the character has different body proportions. 
Our results match the source pose the best and work well on various character shapes. More visual comparisons and animation videos can be found in the supplementary.

Quantitatively, we use the Point-wise Mesh Euclidean Distance (PMD)~\cite{zhou2020unsupervised,wang2020neural} as the evaluation metric. 
% Mixamo provides ground truth data for pose transfer and we can obtain ground truth data of AMASS using SMPL model.
We first evaluate the results on MGN dataset~\cite{bhatnagar2019multi} with all competing methods (the first row of Table.~\ref{table:comparison-pose-transfer}). SAN~\cite{aberman2020skeleton} is not compared because it is trained on Mixamo~\cite{mixamo_2022} and cannot generalize to the unseen skeleton. SPD~\cite{zhou2020unsupervised} trained only on naked human data fails to generalize to clothed human. NBS~\cite{li2021nbs} directly trained on MGC and thus achieves relatively good result. Our full model achieves the best result by using all possible stylized characters.

In addition, we evaluate our method and competing methods on a more challenging dataset Mixamo~\cite{mixamo_2022}, with more stylized test characters. The results are reported in the second row of  Table.~\ref{table:comparison-pose-transfer}. SPD~\cite{zhou2020unsupervised} is not compared since it can only handle meshes with the same mesh connectivity.
All competing methods cannot generalize well to stylized characters in Mixamo and fail significantly in terms of PMD. Our full model results in the lowest PMD which demonstrates the performance of our model on more stylized characters. Our ablation model trained only on AMASS data also scores better than other methods. It shows the generalization of our method when being only trained on limited data.
% We also train our model only on AMASS dataset to further demonstrate the generalization of our model. The results are reported in Table~\ref{table:comparison-pose-transfer}.
% To demonstrate that our model can transfer pose between more stylized characters, we make a comparison with Pinocchio and NBS on Mixamo. SPD can only take the mesh with the same connectivity as input, so it cannot be tested on Mixamo. The results are reported in Table~\ref{table:comparison-mixamo}.

\begin{table}[t]
\centering
\setlength{\tabcolsep}{0.5em}
\begin{tabular}{ccccccc}
\hline
& Pinocchio~\cite{baran2007pinocchio} & SAN~\cite{aberman2020skeleton} & NBS~\cite{li2021nbs} & SPD~\cite{zhou2020unsupervised} & \makecell{ Ours \\  (AMASS)} & \makecell{Ours \\ (full)} \\ 
\hline\hline
\makecell{PMD\ $\downarrow$ \\ on MGN~\cite{bhatnagar2019multi}} 
& 3.145  & - & 1.498 & 5.649 & 2.878   & \textbf{1.197}        \\ \hline
\makecell{PMD\ $\downarrow$ \\ on Mixamo~\cite{mixamo_2022}} 
& 6.139 & 5.581 & 3.875 & - & 3.412  & \textbf{2.393}        \\ \hline
\end{tabular}
% \vspace{+1mm}
\caption{Quantitative comparison of pose transfer results on MGN and Mixamo.}
% \vspace{-4mm}
\label{table:comparison-pose-transfer}
\end{table}

\begin{figure}
    \centering
    \includegraphics[width=\textwidth]{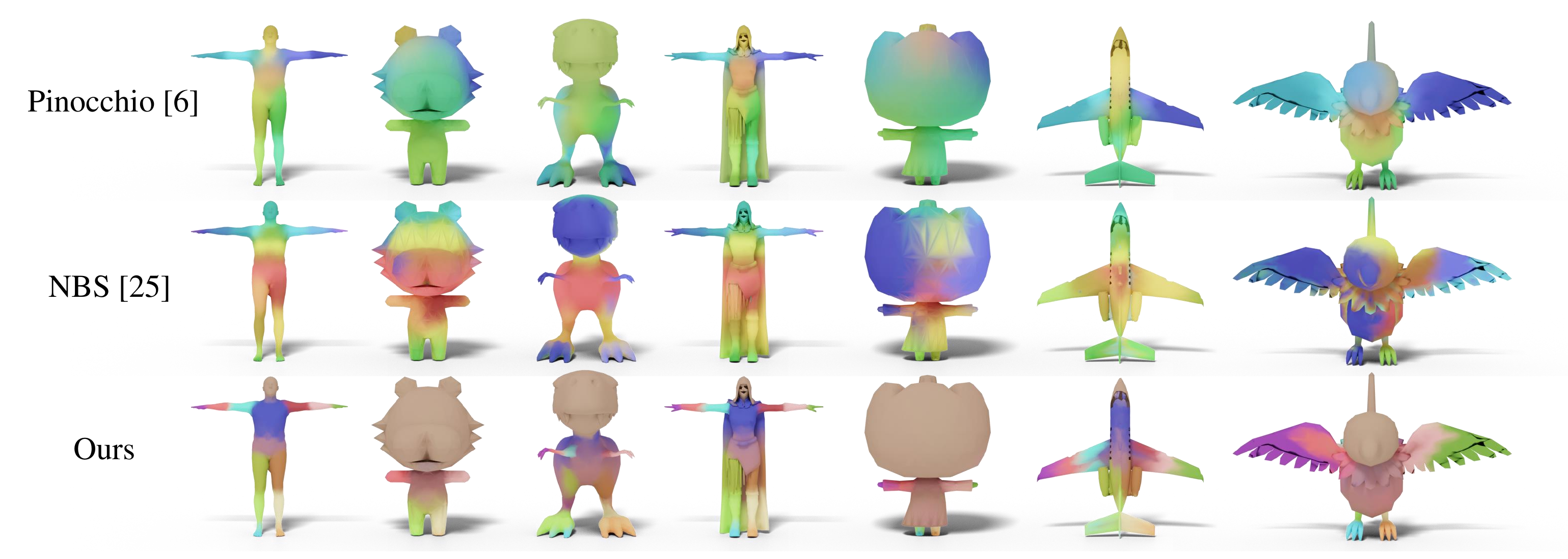}
    \caption{Visualization of deformation parts based on predicted skinning weights from each method (in row). Each part is denoted by a unique color.}
    \label{fig:comparison-consistency}
\end{figure}

% ----------------------semantic consistency----------------------------
\subsection{Deformation Part Semantic Consistency}
Our predicted deformation parts denote the same body region across different characters. Although this can be demonstrated by our pose transfer results, we further validate it by conducting an explicit semantic consistency evaluation. 

\begin{table}
\centering
\setlength{\tabcolsep}{1.2em}
\begin{tabular}{ccccc}
\hline
& Pinocchio~\cite{baran2007pinocchio} & NBS~\cite{li2021nbs} & \makecell{ Ours \\ (AMASS)} & \makecell{Ours \\ (full)} \\ 
\hline\hline
\makecell{Pred $\rightarrow$ GT\ $\uparrow$ \\ on Mixamo~\cite{mixamo_2022}} 
& 0.833    & 0.870 & 0.886    & \textbf{0.993}        \\ \hline
\makecell{GT $\rightarrow$ Pred\ $\uparrow$  \\ on Mixamo~\cite{mixamo_2022}} 
& 0.886    & 0.808 & 0.827    & \textbf{0.947}      \\ \hline
\end{tabular}
% \vspace{+1mm}
\caption{Semantic consistency scores for deformation part prediction. We compare with Pinocchio~\cite{baran2007pinocchio} and NBS~\cite{li2021nbs} on Mixamo~\cite{mixamo_2022} in both correlation directions.}
% \vspace{-4mm}
% Percentage of correct part correlation. Pred $\rightarrow$ GT denotes for each predicted deformation part the percentage that it is correlated to the correct ground truth part. GT $\rightarrow$ Pred denotes for each ground truth deformation part the percentage that it is correlated to the correct predicted part.
\label{table:comparison-consistency}
\end{table}

In Sec.~\ref{sec:training}, we define the vertex belongings to each deformation part by selecting the vertex maximum skinning weight. Therefore, each deformation part can be defined as a mesh semantic part with a group of vertices belonging to it.
Then our goal is to evaluate such semantic part consistency across subjects.
% The deformation part is defined by the skinning weight, which assigns a soft label for each vertex. To make the evaluation easier, we use the deformation part with the greatest skinning weight as the label of a vertex. In this way, we obtain a semantic segmentation of the mesh.
Because existing ground truth annotation, i.e., traditional skeletal skinning weights, cannot be directly used for evaluation, we design an evaluation protocol similar to \cite{fernandez2020unsupervised,jakab2021keypointdeformer} for semantic part consistency. 
More specifically, first, we get our and ground truth mesh semantic parts based on the predicted and ground truth skinning weights respectively. 
Second, we calculate the \textit{correlation} from the semantic parts of prediction to the ones of ground truth: a predicted semantic part is \textit{correlated} with the ground truth part with the most number of overlapped vertices. 
Then we can derive the \textit{``consistency score''} of each predicted part as the maximum percentage of the correlated ground truth parts with the same semantics from all characters in the dataset.
% We calculate the percentage of the most correlated ground truth part out of all characters and treat it as the correct correlation.
The final metric is the average of the consistency score over all predicted parts. 
We denote the above evaluation metric as Pred $\rightarrow$ GT since we find the correlation from predictions to ground truth. GT $\rightarrow$ Pred can be calculated in the opposite direction, i.e., for each ground truth part, we find the most correlated predicted part.

We compare our results with Pinocchio and NBS which rely on a fixed skeleton template and thus can predict skinning weights with the same semantic parts for different characters. The comparison result on Mixamo characters is shown in Table.~\ref{table:comparison-consistency}. 
Our full model achieves the best and close to $1$ correlation accuracy compared to others. Ours trained only on AMASS achieves a similar average performance to comparison methods. We note that NBS used predefined skeleton~\cite{loper2015smpl} for training, while ours is not supervised by any body part labels. 
% \yang{descrition based on table. Our model trained only on AMASS slightly outperforms NBS. However, compared to NBS which uses the pre-defined skeleton from SMPL, we do not use any body part label during training.}

We also visualize the skinning weights predicted from ours and comparison methods in Fig.~\ref{fig:comparison-consistency}. For each method, we used consistent color for deformation parts to reflect the same semantic. Our skinning paints characters consistently on semantic regions while the comparison methods fail on some body parts.

\subsection{Ablation Study}
We conduct ablation studies on Mixamo dataset to investigate the effectiveness of each component. \textbf{w/o edge loss} is trained without the edge length constraint. \textbf{w/o pseudo} is trained without cycle loss from the pseudo-ground truth. \textbf{w/o skinning} is trained with out skinning weight loss. Table.~\ref{table:ablation} shows the evaluation results on Mixamo data. Our full model achieves the best performance.

\begin{table}
\centering
\setlength{\tabcolsep}{0.6em}
\begin{tabular}{ccccc}
\hline
 & w/o edge loss & w/o pseudo  & w/o skinning loss & Ours (full) \\ \hline\hline
\makecell{PMD\ $\downarrow$ \\ on Mixamo~\cite{mixamo_2022}} 
& 2.450 & 2.601 & 2.978  & \textbf{2.393}        \\ \hline
\end{tabular}
% \vspace{+1.5mm}
\caption{Quantitative evaluation results on ablation methods.}
% \vspace{-4mm}
\label{table:ablation}
\end{table}

% & \makecell{ w/o skinning  \\ consistency} = 2.486

% \subsection{Robustness to Mesh Connectivity and Rest Pose}
% Optional.

% \subsection{Applications}

% \vspace{-5mm}
\section{Conclusion and Future Work}
We present a novel learning-based framework to automatically transfer poses between stylized 3D characters without skeletal rigging. Our model can handle characters with diverse shapes, topologies, and mesh connectivities. We achieve this by representing the characters in a unified articulation model and predicting the deformation skinning and transformations when given the desired pose. 
Our model can utilize all types of existing character data, e.g., with motion or static, and thus can have great generalization to various unseen stylized characters. 

\subsubsection{Limitations and Future Work.}
Our model focuses on pose transfer and is not optimized for motion transfer in the temporal domain. Jittering and penetration problems~\cite{villegas2021contact} may occur when using our proposed method for animation. We apply the edge length constraint to prevent the mesh from breaking but no other explicit controls are involved. Data-driven deformation constraints~\cite{gao2019sparse} and better geometric regularizations~\cite{sorkine2007rigid} could prevent the implausible deformations further. Our current framework requires the rest pose mesh as an additional input. Canonicalization methods~\cite{musoni2021reposing} might be helpful to ease this requirement.
We are looking for automating the process of the character pose transfer, yet in real content authoring scenarios, user input is still desired to increase tool usability and content diversity. We look forward to future endeavors on expressive pose transfer animations with intuitive user controls.
% \begin{enumerate}
%     \item Do not support motion transfer. Only pose transfer.
%     \item No explicit method to guarantee different deformation parts won't break down after transformation. May add more constraint (e.g., bones)
%     \item Need rest pose as input. Can add canonicalization in the future.
%     \item Enable user input to specify the correspondence for characters with ambiguity
% \end{enumerate}

\subsubsection{Acknowledgement.}
This work is funded by a gift from Adobe Research and the Deutsche Forschungsgemeinschaft (DFG, German Research Foundation) - 409792180 (Emmy Noether Programme, project: Real Virtual Humans). Gerard Pons-Moll is a member of the Machine Learning Cluster of Excellence, EXC number 2064/1 – Project number 390727645.

\bibliographystyle{splncs04}
\bibliography{egbib}
\end{document}